\documentclass[conference,a4paper]{IEEEtran}
\IEEEoverridecommandlockouts

\usepackage[hidelinks]{hyperref}
\usepackage[cmex10]{amsmath}%American Math Society(AMS) math formatting
\usepackage{amssymb,amsfonts}%AMS extra symbols and fonts
\usepackage{dblfloatfix}%fix double column figure ordering and placement

\usepackage[ruled,vlined]{algorithm2e}
\usepackage{graphicx}
\graphicspath{{Figures/PDF/}{Figures/PNG/}}

\usepackage{booktabs}
\usepackage{siunitx}
\usepackage[numbers,compress]{natbib}
\usepackage{texnames}
\usepackage{bm,bbm}
\usepackage{orcidlink}
\usepackage{multirow}

\begin{document}

\title{\uppercase{Deep S2P: Integrating Learning Based Stereo Matching into the Satellite Stereo Pipeline}
}

% TODO: anonymize authors!
%\author{	\IEEEauthorblockN{}
%	\IEEEauthorblockA{\textit{Victoria University of Wellington}\\
%		6140 Wellington, New Zealand\\
%		alejandro.frery@vuw.ac.nz}
%	\and
%	\IEEEauthorblockN{Hui Zhang\orcidlink{0000-0002-5283-7350}}
%	\IEEEauthorblockA{\textit{Inner Mongolia University}\\
%		010021 Hohhot, China\\
%		hui.zhang@imu.edu.cn}
%	\and
%	\IEEEauthorblockN{Andrea Rey\orcidlink{0000-0002-9185-1382}}
%	\IEEEauthorblockA{\textit{Universidad Nacional de Hurlingham}\\
%		1688 República Argentina\\
%		andrea.rey@unahur.edu.ar}
%}

\author{El\'ias Masquil$^1$ \hfill Thibaud Ehret$^2$ \hfill Pablo Musé$^{1,3}$ \hfill Gabriele Facciolo$^{3,4}$\\
$^1$ IIE, Facultad de Ingeniería, Universidad de la República, Uruguay \quad\quad $^2$ AMIAD, Pôle Recherche, France\\
$^3$ Université Paris-Saclay, ENS Paris-Saclay, CNRS, Centre Borelli, 91190, Gif-sur-Yvette, France
\\
$^4$ Institut Universitaire de France 

\vspace{-1em}
}

%\def\tenpt{\def\baselinestretch{1.1}\let\normalsize\small\normalsize}
%\tenpt

\maketitle
\begin{abstract}

% Adjust line of writing (benchmark saturation and quality perception) if most deep stereo networks aren't that good

Digital Surface Model generation from satellite imagery is a core task in Earth observation and is commonly addressed using classical stereoscopic matching algorithms in satellite pipelines as in the Satellite Stereo Pipeline (S2P). While recent learning-based stereo matchers achieve state-of-the-art performance on standard benchmarks, their integration into operational satellite pipelines remains challenging due to differences in viewing geometry and disparity assumptions. In this work, we integrate several modern learning-based stereo matchers, including StereoAnywhere, MonSter, Foundation Stereo, and a satellite fine-tuned variant of MonSter, into the Satellite Stereo Pipeline, adapting the rectification stage to enforce compatible disparity polarity and range. We release the corresponding code to enable reproducible use of these methods in large-scale Earth observation workflows. Experiments on satellite imagery show consistent improvements over classical cost-volume-based approaches in terms of Digital Surface Model accuracy, although commonly used metrics such as mean absolute error exhibit saturation effects. Qualitative results reveal substantially improved geometric detail and sharper structures, highlighting the need for evaluation strategies that better reflect perceptual and structural fidelity. At the same time, performance over challenging surface types such as vegetation remains limited across all evaluated models, indicating open challenges for learning-based stereo in natural environments.
\end{abstract}


\section{Introduction}
\label{sec:introduction}

The generation of Digital Surface Models (DSMs) from satellite and aerial imagery is a central task in Earth observation (EO). These 3D products are necessary for applications ranging from urban planning and infrastructure monitoring to disaster response and environmental studies. For decades, classical photogrammetric pipelines, most notably Semi-Global Matching (SGM)~\cite{hirschmuller2007stereo} and open-source tools such as the Satellite Stereo Pipeline (S2P)~\cite{amadei2025s2p} and the Ames Stereo Pipeline (ASP)~\cite{asp}, have been the standard for stereo reconstruction. Their popularity stems from their computational efficiency, scalability, and robustness.

In recent years, deep learning has reshaped the landscape of stereo matching. End-to-end architectures such as PSMNet~\cite{chang2018pyramid} and RAFT-Stereo~\cite{lipson2021raft}, often referred to as advanced correlators, have consistently established new state-of-the-art results on standard benchmarks. When adapted to EO imagery, these methods have demonstrated clear improvements in reconstruction accuracy compared to classical pipelines. 

Despite these advances, integrating deep learning-based correlators into operational EO stereo pipelines remains non-trivial. Most learning-based stereo models are trained on close-range datasets that differ substantially from satellite imagery in terms of viewing geometry and epipolar geometry. In particular, these models implicitly assume disparity distributions that are consistent with their training data. Naively replacing classical correlators within existing pipelines can result in sub-optimal performance or unstable predictions when the expected disparity magnitude or sign is violated~\cite{mari2022disparity, masquil2026diachronic}.

Our contribution is the integration of state-of-the-art learning-based stereo matchers into the Satellite Stereo Pipeline (S2P)~\cite{de2014automatic,amadei2025s2p}. We extend S2P to support advanced correlators, including StereoAnywhere~\cite{bartolomei2025stereo}, MonSter~\cite{cheng2025monster}, Foundation Stereo~\cite{wen2025foundationstereo}, and an EO fine-tuned variant~\cite{masquil2026diachronic} of MonSter. We adapt the rectification stage to ensure compatibility with their disparity assumptions and account for the different sign conventions in the outputs of the different models. We validate this integration through a comprehensive experimental evaluation on satellite imagery.

Across our experiments, learning-based matchers consistently outperform traditional cost-volume-based methods. Quantitatively, the improvements are reflected in lower DSM errors under standard evaluation metrics. However, we observe that commonly used metrics, such as the mean absolute error (MAE), tend to saturate beyond a certain performance level: although learning-based methods achieve higher scores, the numerical gains remain relatively modest.

In contrast, qualitative inspection of the reconstructed DSMs reveals a marked improvement in perceptual quality, with sharper structures, finer geometric details, and improved delineation of complex objects, particularly in urban environments. This discrepancy suggests that existing evaluation metrics may not fully capture the benefits of modern stereo matchers. It also raises the possibility that current ground-truth DSMs, limited by their own spatial resolution and noise characteristics, may restrict measurable improvements. These observations motivate the exploration of evaluation metrics and assessment strategies better aligned with geometric fidelity and structural detail in high-quality DSM reconstruction.

Despite the overall performance gains, the improvements are not consistent across all surface types. Vegetated areas, in particular, remain challenging, with all evaluated models showing larger reconstruction errors than in man-made or ground surfaces. Addressing vegetation remains an open challenge for learning-based stereo in Earth observation.

\section{Related Work}
\label{sec:related_work}

\smallskip\noindent\textbf{Classical stereo pipelines in EO}
For decades, stereo reconstruction in Earth observation (EO) has relied on classical methods such as Semi-Global Matching (SGM)~\cite{hirschmuller2007stereo} and More Global Matching (MGM)~\cite{facciolo2015mgm}, valued for their balance of accuracy and efficiency. Open-source pipelines like  S2P~\cite{de2014automatic,amadei2025s2p} and the Ames Stereo Pipeline (ASP)~\cite{asp} have established themselves as standard tools, largely thanks to their ability to process arbitrarily large images through explicit tiling and seamless mosaic stitching. These pipelines remain the backbone of operational DSM and DEM generation at scale.

\smallskip\noindent\textbf{Deep learning correlators}
Recent years have seen a shift from hand-crafted matching costs to end-to-end deep learning approaches for stereo reconstruction. Early methods such as PSMNet~\cite{chang2018pyramid} and GANet~\cite{wang2022keypoint} introduced 3D cost-volume aggregation, while iterative optimization-based architectures like RAFT-Stereo~\cite{lipson2021raft} and its successors~\cite{xu2023iterative} have achieved state-of-the-art accuracy through recurrent refinement. Building on these ideas, more recent models such as MonSter~\cite{cheng2025monster}, StereoAnywhere \cite{bartolomei2025stereo}, FoundationStereo \cite{wen2025foundationstereo} trained predominantly on synthetic data and leveraging monocular priors, have shown strong generalization in diverse real-world settings.

 When adapted to EO data, these advanced correlators have consistently outperformed SGM and other classical baselines on benchmarks such as WHU-Stereo~\cite{li2023whu}, demonstrating their potential for large-scale satellite stereo reconstruction. Recently,~\cite{masquil2026diachronic} introduced a fine-tuned variant of MonSter trained on satellite imagery, explicitly addressing both same-date and diachronic stereo pairs.% At the same time, their generalization across geographic domains remains uneven, and their deployment on continental or global scales continues to raise practical challenges.

% \smallskip\noindent\textbf{Tiling in deep methods.}
% Because satellite images far exceed GPU memory limits, tiling is unavoidable in practice. Classical pipelines are designed around this constraint: they operate locally and then explicitly stitch tiles, producing globally consistent outputs. Deep networks, however, depend on spatial context within their receptive fields, and this assumption is violated at tile boundaries. As a result, objects that cross tile edges can lead to systematic artifacts in the final DSM \cite{huang2018tiling}. Similar problems have been studied in dense prediction tasks such as semantic segmentation, where strategies like “Flip-n-Slide” have been proposed to mitigate boundary effects \cite{abrahams2024concise}. To date, however, such solutions have rarely been explored in stereo matching, where the problem is compounded by the complexity of cost-volume inference.

\smallskip\noindent\textbf{Vegetation and natural environments}
Another persistent difficulty for stereo reconstruction lies in vegetated areas. Forests and agricultural fields exhibit a combination of repetitive patterns, textureless regions, and non-rigid canopy structures, making correspondence highly ambiguous. Seasonal variation and wind-induced motion further exacerbate the problem, leading to noisy or unreliable surface models. These challenges are well-documented not only in EO benchmarks but also in forestry and agricultural robotics~\cite{karim2024review,dandrifosse2020imaging}. While the reconstructions of urban scenes tend to be reliable, vegetation remains a weak spot for modern correlators. Specialized multi-view stereo methods have been proposed for forests and plants~\cite{liu2024cdp,chen2025fs}, but such domain-specific adaptations are not yet common in EO. Addressing this gap is essential for deep learning methods to move from controlled benchmarks toward reliable global-scale DSM generation.

% best pairs + s2p-hd table
\begin{table*}[t]
\centering
\caption{Performance comparison on stereo pairs from the GRSS dataset. Best results are shown in \textbf{bold}, second best are \underline{underlined}.}
%\resizebox{0.6\linewidth}{!}{
\label{tab:grss_bestpairs}
\begin{tabular}{@{}lccccc@{}}
\toprule
\textbf{Method} & \textbf{P90} & \textbf{NMAD} & \textbf{RMSE} & \textbf{MAE} & \textbf{\% Valid Points} \\
\midrule
S2P-HD (SGM) 
& $2.30\pm1.86$ 
& $1.03\pm0.35$ 
& $5.19\pm4.16$ 
& $2.25\pm0.87$ 
& $78.51\pm19.57$ \\

S2P-HD (MGM) 
& $2.82\pm3.91$ 
& $1.13\pm0.74$ 
& $6.15\pm6.47$ 
& $3.07\pm2.86$ 
& $81.15\pm15.44$ \\
\midrule
Diachronic Stereo 
& $3.22\pm2.60$ 
& $1.21\pm0.95$ 
& $3.66\pm2.11$ 
& $2.28\pm1.22$ 
& \bm{$90.89\pm19.29$} \\

FoundationStereo 
& \bm{$1.91\pm2.23$} 
& \bm{$0.53\pm0.26$} 
& \bm{$3.32\pm1.67$} 
& \bm{$1.96\pm0.92$} 
& $88.51\pm8.80$ \\

MonSter 
& \underline{$2.06\pm2.20$} 
& \underline{$0.69\pm0.33$} 
& \underline{$3.37\pm1.74$} 
& \underline{$2.05\pm0.93$} 
& $87.58\pm18.44$ \\

StereoAnywhere 
& $2.70\pm2.70$ 
& $0.99\pm0.63$ 
& $3.59\pm1.79$ 
& $2.20\pm1.02$ 
& \underline{$90.45\pm12.73$} \\
\bottomrule
\end{tabular}
%}
\vspace{-1.5em}
\end{table*}

\section{Methodology}
\label{sec:method}

Deep S2P extends the Satellite Stereo Pipeline (S2P) by replacing its classical stereo matching stage with modern learning-based stereo matchers. While S2P already provides a robust and scalable rectification and matching framework for satellite imagery, learning-based matchers impose additional geometric constraints on the rectified image pairs. In particular, they assume a consistent disparity polarity and operate reliably only within a bounded disparity range. 

S2P rectifies satellite image pairs using RPC-based virtual correspondences following the approach of~\cite{de2014automatic}, optionally refined with a horizontal shear to reduce the disparity range and a horizontal translation estimated from sparse image matches. In Deep S2P, we reuse the standard S2P rectification pipeline and introduce lightweight adjustments to ensure compatibility with learning-based stereo matchers.

\smallskip\noindent\textbf{Disparity polarity and altitude consistency}
Most learning-based stereo matchers implicitly assume unipolar disparities, meaning that valid correspondences are shifted horizontally in a single direction across the image pair and that disparity magnitude increases with scene altitude. In contrast, standard satellite rectification may produce mixed-sign disparities depending on the viewing geometry and image ordering.

To enforce unipolarity, we apply a global horizontal translation to the rectified right image such that all observed disparities share the same sign. Concretely, given a set of sparse matches after initial rectification, we estimate the minimum horizontal displacement and shift the right image accordingly so that all disparities are strictly positive or strictly negative in rectified coordinates.

After enforcing unipolarity, we verify altitude consistency by projecting the center of the region of interest at two different altitudes using the RPC models and rectifying homographies. If the measured disparity does not increase with altitude, the rectification is recomputed using the opposite disparity sign configuration in the horizontal registration step. This procedure ensures that higher elevations correspond to larger disparity values, matching the geometric assumptions of learning-based stereo models.

We follow Algorithm 1 from~\cite{masquil2026diachronic} for this polarity and altitude consistency enforcement, with two simplifications tailored to the present setting. First, we do not apply residual vertical correction, as same-date stereo pairs typically exhibit limited vertical misalignment. Second, we rely on SIFT matches~\cite{lowe2004sift} for sparse correspondence estimation, which are sufficient and robust in the standard same-date satellite stereo scenario considered in this work.

\smallskip\noindent\textbf{Disparity range handling}
Beyond polarity, learning-based stereo matchers are sensitive to the magnitude of disparities. Empirically, we observe that operating near zero disparity leads to unstable predictions, likely due to mismatches with the disparity statistics of standard training datasets. To mitigate this effect, after enforcing unipolarity, we apply an additional horizontal translation to the right image that shifts the entire disparity range away from zero. In all experiments, we use a minimum disparity margin of 50 pixels. 

\smallskip\noindent\textbf{Model specific conventions}
Different learning-based matchers adopt different disparity sign conventions. Deep S2P standardizes all outputs to the S2P convention through model-specific post-processing. For example, StereoAnywhere and MonSter produce disparities with opposite sign conventions. We apply the appropriate sign correction to each model output.

\smallskip\noindent\textbf{Left-right consistency filtering}
To improve robustness and suppress unreliable estimates, Deep S2P applies a left-right consistency check to the predicted disparities. We compute both left-to-right and right-to-left disparity maps and invalidate pixels for which disparities are inconsistent by more than 2 pixels. Disparities failing this consistency check are discarded from subsequent DSM generation and evaluation.

\section{Results}

% semantic best pairs table
\begin{table*}[t]
\centering
\caption{Semantic-wise DSM error analysis on the GRSS dataset (best pairs). Best results are shown in \textbf{bold}, second best are \underline{underlined}.}
\label{tab:grss_bestpairs_semantic}
\resizebox{0.72\linewidth}{!}{
\begin{tabular}{@{}llcccc@{}}
\toprule
\textbf{Method} & \textbf{Class} & \textbf{P90} & \textbf{NMAD} & \textbf{RMSE} & \textbf{MAE} \\
\midrule
% --- Ground (unchanged) ---
S2P-HD (SGM) & Ground 
& $1.35\pm1.30$ & $0.83\pm0.35$ & $4.10\pm4.69$ & $1.85\pm0.90$ \\
S2P-HD (MGM) & Ground 
& $1.44\pm1.32$ & $0.81\pm0.38$ & $4.01\pm3.43$ & $2.04\pm1.08$ \\
Diachronic Stereo & Ground 
& $1.27\pm2.00$ & $0.85\pm0.62$ & $2.44\pm1.49$ & $1.67\pm0.97$ \\
FoundationStereo & Ground 
& \bm{$0.78\pm1.33$} & \bm{$0.40\pm0.15$} & \bm{$2.37\pm1.36$} & \bm{$1.52\pm0.79$} \\
MonSter & Ground 
& \underline{$0.79\pm1.37$} & \underline{$0.48\pm0.23$} & \underline{$2.40\pm1.47$} & \underline{$1.61\pm0.94$} \\
StereoAnywhere & Ground 
& $1.07\pm1.41$ & $0.65\pm0.23$ & $2.52\pm1.40$ & $1.65\pm0.75$ \\

\midrule
% --- Roof (unchanged) ---
S2P-HD (SGM) & Roof 
& $2.08\pm1.67$ & $0.88\pm0.35$ & $4.08\pm5.86$ & $2.13\pm2.20$ \\
S2P-HD (MGM) & Roof 
& $2.00\pm1.63$ & $0.83\pm0.37$ & $3.12\pm2.64$ & $1.91\pm1.23$ \\
Diachronic Stereo & Roof 
& $2.24\pm2.56$ & $1.11\pm1.06$ & $2.95\pm3.15$ & $1.95\pm1.70$ \\
FoundationStereo & Roof 
& \bm{$1.28\pm1.63$} & \bm{$0.50\pm0.31$} & \bm{$2.53\pm2.42$} & \bm{$1.66\pm1.33$} \\
MonSter & Roof 
& \underline{$1.29\pm1.68$} & \underline{$0.55\pm0.30$} & $2.64\pm2.68$ & \underline{$1.70\pm1.38$} \\
StereoAnywhere & Roof 
& $1.56\pm1.71$ & $0.78\pm0.46$ & \underline{$2.61\pm2.30$} & $1.75\pm1.27$ \\

\midrule
% --- Tree (FIXED highlights) ---
S2P-HD (SGM) & Tree 
& \underline{$6.72\pm3.68$} & \underline{$2.75\pm1.13$} & $5.59\pm3.26$ & \underline{$3.46\pm1.06$} \\

S2P-HD (MGM) & Tree 
& \bm{$6.44\pm3.73$} & \bm{$2.62\pm1.20$} & $5.15\pm1.83$ & \bm{$3.39\pm1.09$} \\

Diachronic Stereo & Tree 
& $7.53\pm3.28$ & $2.94\pm0.89$ & \bm{$4.95\pm1.80$} & $3.67\pm1.45$ \\

FoundationStereo & Tree 
& $7.60\pm3.73$ & $3.03\pm1.22$ & $5.05\pm1.56$ & $3.59\pm1.09$ \\

MonSter & Tree 
& $7.53\pm3.71$ & $3.12\pm1.26$ & \underline{$5.01\pm1.57$} & $3.60\pm1.11$ \\

StereoAnywhere & Tree 
& $8.32\pm4.03$ & $3.45\pm1.17$ & $5.39\pm2.00$ & $3.93\pm1.60$ \\

\midrule
% --- All Except Vegetation and Water (unchanged) ---
S2P-HD (SGM) & All Except Vegetation and Water 
& -- & -- & $4.05\pm4.05$ & $1.83\pm0.87$ \\
S2P-HD (MGM) & All Except Vegetation and Water 
& -- & -- & $4.22\pm3.59$ & $2.06\pm1.13$ \\
Diachronic Stereo & All Except Vegetation and Water 
& -- & -- & $2.70\pm2.03$ & $1.73\pm1.05$ \\
FoundationStereo & All Except Vegetation and Water 
& -- & -- & \bm{$2.51\pm1.68$} & \bm{$1.55\pm0.91$} \\
MonSter & All Except Vegetation and Water 
& -- & -- & \underline{$2.57\pm1.82$} & \underline{$1.64\pm1.03$} \\
StereoAnywhere & All Except Vegetation and Water 
& -- & -- & $2.64\pm1.70$ & $1.69\pm0.90$ \\
\bottomrule
\end{tabular}
}
\vspace{-1.5em}
\end{table*}

% worst pairs table
\begin{table*}[t]
\centering
\caption{Performance on challenging stereo pairs from the GRSS dataset. Best results are shown in \textbf{bold}, second best are \underline{underlined}.}
\label{tab:grss_worstpairs}
%\resizebox{0.6\linewidth}{!}{
\begin{tabular}{@{}lccccc@{}}
\toprule
\textbf{Method} & \textbf{P90} & \textbf{NMAD} & \textbf{RMSE} & \textbf{MAE} & \textbf{\% Valid Points} \\
\midrule
Diachronic Stereo
& \underline{$2.10\pm2.42$}
& \underline{$0.44\pm0.59$}
& \bm{$2.63\pm1.66$}
& \bm{$1.24\pm1.20$}
& \bm{$65.72\pm21.77$} \\

FoundationStereo
& \bm{$2.03\pm2.43$}
& \bm{$0.39\pm0.60$}
& \underline{$2.85\pm1.91$}
& \underline{$1.44\pm1.33$}
& \underline{$64.85\pm17.87$} \\

%MonSter
%& \underline{$2.10\pm2.42$}
%& \underline{$0.44\pm0.59$}
%& \bm{$2.63\pm1.66$}
%& \bm{$1.24\pm1.20$}
%& \bm{$65.72\pm21.77$} \\
MonSter
& $4.19\pm4.25$ 
& $0.53\pm0.78$
& $3.75\pm2.10$
& $1.89\pm1.51$
& $61.41\pm26.87$\\

StereoAnywhere
& $3.43\pm4.49$
& $0.63\pm1.10$
& $3.60\pm2.96$
& $1.81\pm2.23$
& $61.13\pm15.48$ \\
\bottomrule
\end{tabular}
%}
\vspace{-1em}
\end{table*}

% Figure
\begin{figure*}[t]
\centering
\setlength{\tabcolsep}{1.5pt}
\renewcommand{\arraystretch}{0.0}
\resizebox{0.9\linewidth}{!}{
\begin{tabular}{ccccccc}
\footnotesize S2P-HD (SGM) &
\footnotesize S2P-HD (MGM) &
\footnotesize StereoAnywhere &
\footnotesize MonSter &
\footnotesize FoundationStereo &
\footnotesize Diachronic &
\footnotesize GT \\[0.15cm]

\includegraphics[width=0.14\textwidth]{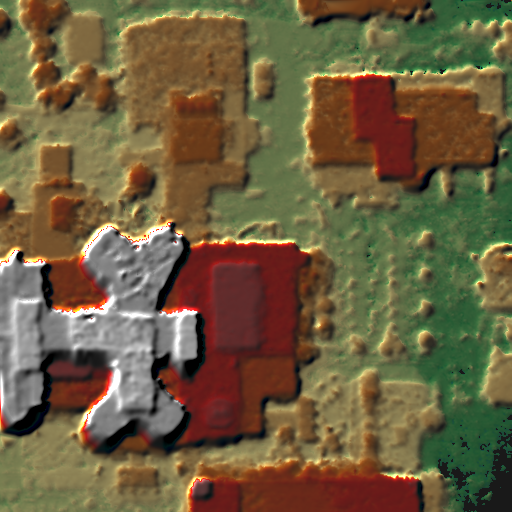} &
\includegraphics[width=0.14\textwidth]{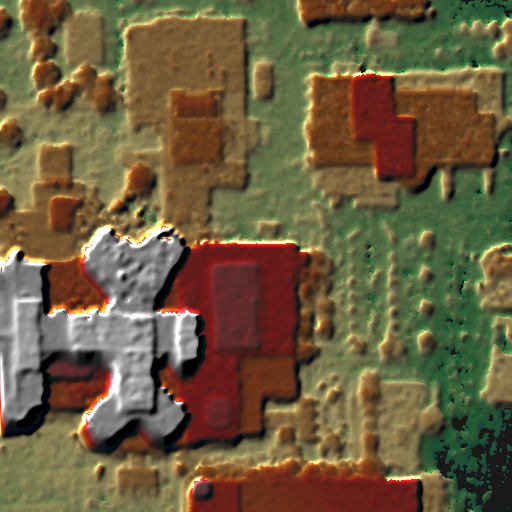} &
\includegraphics[width=0.14\textwidth]{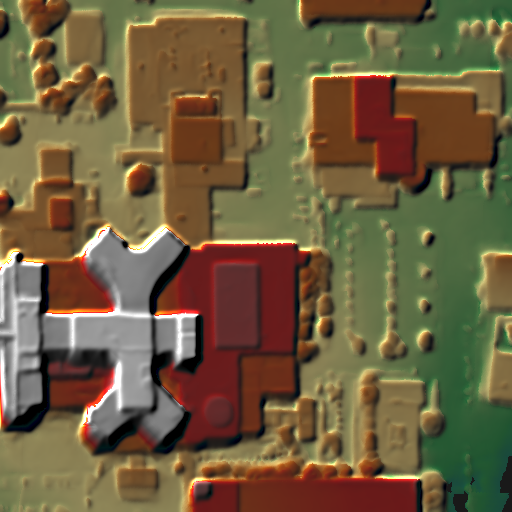} &
\includegraphics[width=0.14\textwidth]{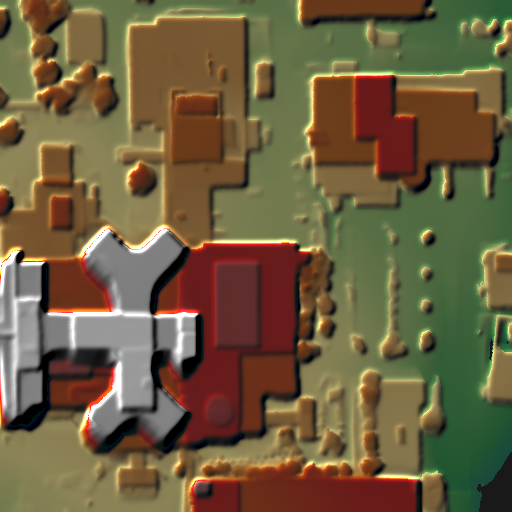} &
\includegraphics[width=0.14\textwidth]{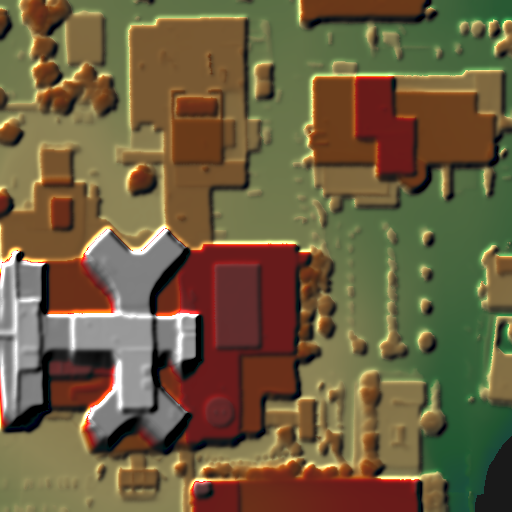} &
\includegraphics[width=0.14\textwidth]{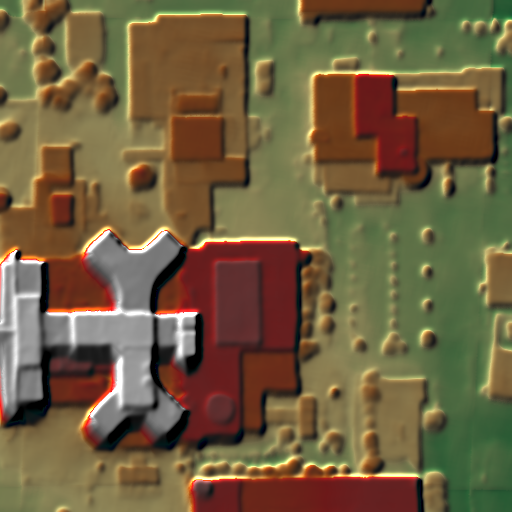} &
\includegraphics[width=0.14\textwidth]{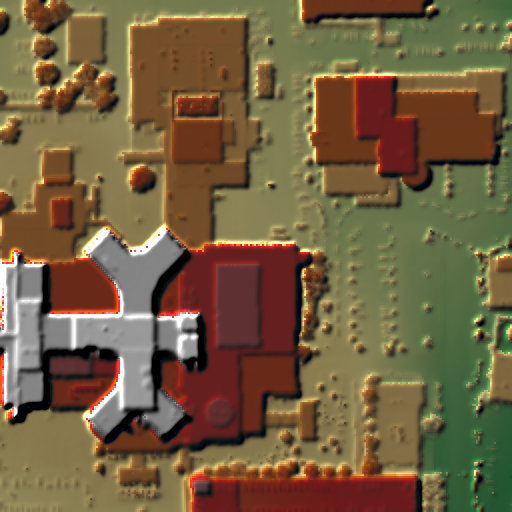} \\[0.05cm]

\includegraphics[width=0.14\textwidth]{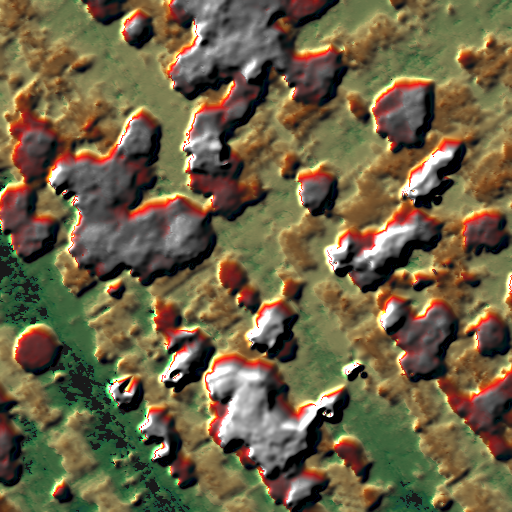} &
\includegraphics[width=0.14\textwidth]{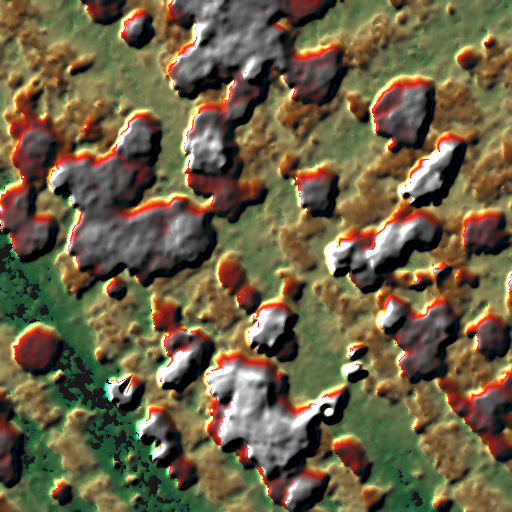} &
\includegraphics[width=0.14\textwidth]{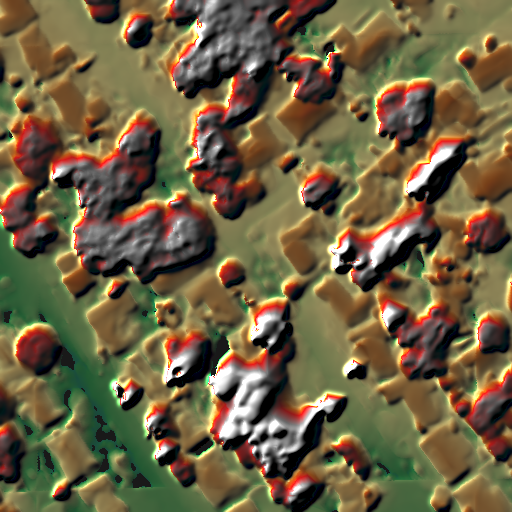} &
\includegraphics[width=0.14\textwidth]{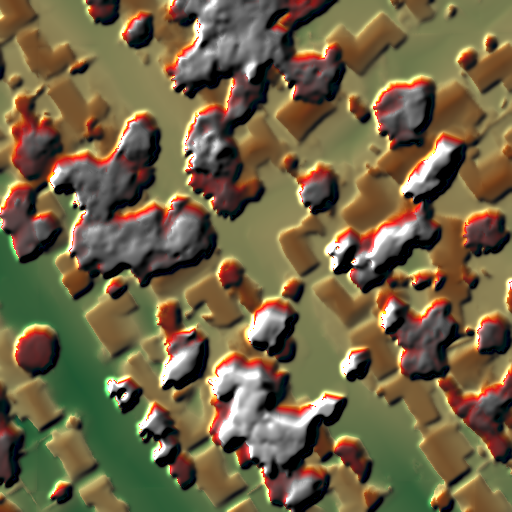} &
\includegraphics[width=0.14\textwidth]{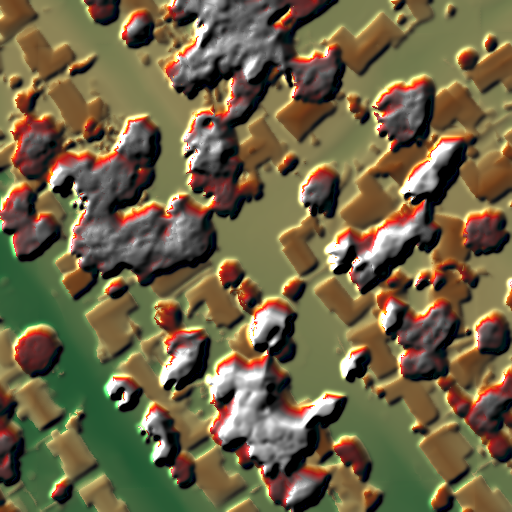} &
\includegraphics[width=0.14\textwidth]{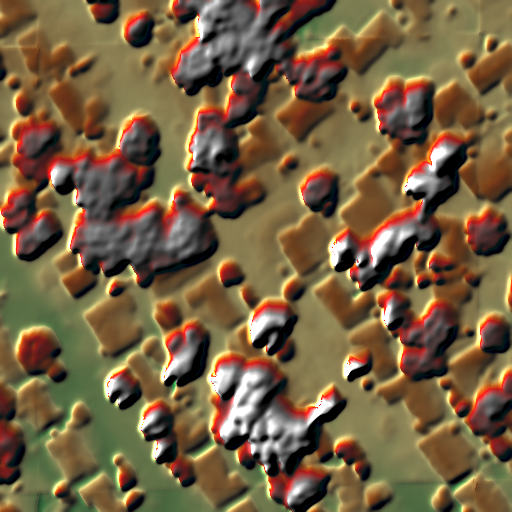} &
\includegraphics[width=0.14\textwidth]{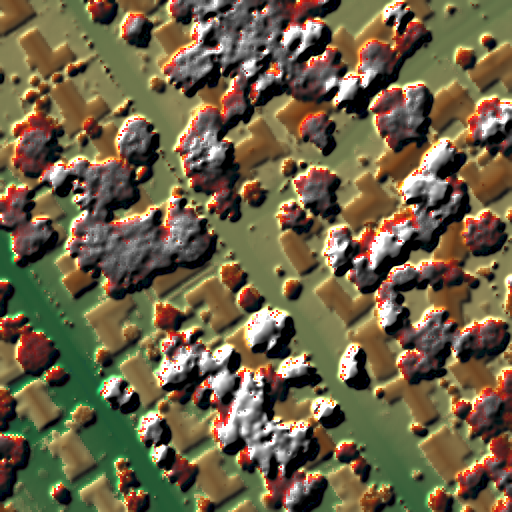}
\end{tabular}
}
\vspace{0.2em}
\caption{DSM reconstructions for two AOIs. Top row: JAX\_068 (urban scene). Bottom row: JAX\_559 (suburban scene with dense vegetation).}
\label{fig:qualitative_dsm}
\vspace{-1.5em}
\end{figure*}

All the experiments were run on an Intel Xeon Gold 6426Y CPU with 32 cores (64 threads) and 252GB of RAM, and an NVIDIA RTX 6000 Ada Generation GPU with 48GB of VRAM. We evaluate Deep S2P on same-date stereo pairs from the 2019 IEEE GRSS Data Fusion Contest dataset \cite{bosch2019semantic}. This dataset consists of multi-view satellite imagery from WorldView-3 over 53 AOIs in Jacksonville, Florida. As a first benchmark, we consider the pairs identified in the evaluation protocol of \cite{amadei2025s2p}, which represent favorable acquisition geometry. On this subset, we compare the learning based correlators integrated into Deep S2P against the top-performing classical correlators available in s2p-hd: MGM \cite{facciolo2015mgm} and SGM \cite{hirschmuller2007stereo, fixstars_libsgm_2023}, as reported by the authors \cite{amadei2025s2p}. In addition to global DSM accuracy, we perform a semantic-aware evaluation by analyzing errors separately over different surface classes, such as building roofs, vegetation, ground, etc. To further asses robustness, we conduct a complementary evaluation on challenging stereo pairs that violate the heuristic selection criteria proposed in \cite{facciolo2017automatic}, which exhibit challenging viewing angles between them.

We use xDEM~\cite{xdem_2024} for DSM alignment and evaluation, excluding points only when invalidated by the left–right consistency check described in Section~\ref{sec:method}. To compensate for residual biases due to RPC inaccuracies, we compute the median altitude error for each method across all scenes and apply a corresponding global vertical offset to the reconstructed DSMs prior to evaluation. Performance is assessed using  90th Percentile Error, Normalized Median
Absolute Deviation (NMAD) \cite{amadei2025s2p}, Root Mean Squared Error (RMSE), Mean Absolute Error (MAE), and Percentage of Valid Points. We report the average $\pm$ standard deviation across all AOIs.

\smallskip\noindent\textbf{2019 IEEE GRSS Data Fusion Contest - Best Pairs}
We take same-date stereo pairs and filter them to have baseline angles between 5° and 45°, incidence angles below 40°, and minimal acquisition time differences (averaging 11 seconds). The evaluation set consists of one pair of images per AOI and includes LiDAR-derived ground-truth DSMs at 50 cm resolution with semantic annotations.

Table~\ref{tab:grss_bestpairs} reports the performance of the learning based correlators integrated into Deep S2P and the S2P-HD baseline. All learning based methods consistently outperform the classical baselines across all error metrics, at the cost of approximately a twofold increase in runtime. Beyond improved accuracy, learning based correlators also produce substantially more complete DSMs, with a higher proportion of valid reconstructed points. Among the evaluated models, FoundationStereo achieves the strongest overall performance.% We note that the s2p-hd results reported here differ from those in~\cite{amadei2025s2p}, which we attribute to differences in the evaluation protocol, as our evaluation is performed over all valid points and does not apply the same outlier filtering used by the authors.

Figure~\ref{fig:qualitative_dsm} shows a qualitative comparison of the reconstructed DSMs. Learning-based methods consistently produce sharper structures and finer geometric detail than classical correlators. While the quantitative improvements are sometimes modest, the visual differences are clearly apparent, including for Diachronic Stereo, which shows limited numerical gains but noticeably improved structural detail.

Table~\ref{tab:grss_bestpairs_semantic} reports class-wise performance across surface types. All methods achieve their lowest errors on man-made and ground surfaces, while errors increase substantially over vegetation. Notably, this is also observed for classical correlators, but the performance gap between learning-based and classical methods narrows for this class. In addition to the inherent difficulty of reconstructing vegetation, this effect may be further amplified by residual temporal inconsistencies between the satellite imagery and the ground-truth DSM, to which vegetation is especially sensitive. This behavior is visible in the bottom row of Figure~\ref{fig:qualitative_dsm}, where dense vegetation leads to irregular and blobby surface geometry across all methods, a difficulty that is also apparent in the ground-truth DSM.

\smallskip\noindent\textbf{2019 IEEE GRSS Data Fusion Contest - Worst Pairs}
Additionally we define a set of geometrically challenging stereo pairs. For each AOI, we select the same-date pair with the closest acquisition time that does not satisfy the geometric filtering criteria above. These pairs might have either excessively small or large baselines, or extreme incidence angles.

Under challenging viewing geometry, DSM completeness becomes the dominant limitation. As shown in Table~\ref{tab:grss_worstpairs}, all learning-based correlators produce a large fraction of missing points, with only about 61–66\% of points reconstructed. When restricting evaluation to valid pixels, however, the reconstruction errors remain relatively low. In this setting, Diachronic Stereo achieve the most robust overall performance, while FoundationStereo exhibits comparable results and StereoAnywhere shows a more pronounced degradation in both accuracy and completeness.

\section{Discussion}
We have presented Deep S2P, an extension of S2P that integrates modern learning-based stereo matchers into operational satellite workflows. By adapting the rectification stage to enforce compatible disparity polarity and range constraints, we enable the use of StereoAnywhere, MonSter, FoundationStereo, and Diachronic Stereo within the S2P pipeline.
Experiments on the GRSS 2019 dataset show that learning-based methods consistently outperform classical approaches in both accuracy and completeness. FoundationStereo achieves the best quantitative results, though all deep learning models produce visibly sharper structures and finer geometric detail than SGM or MGM. The modest numerical gains despite clear qualitative improvements suggest both that standard metrics inadequately capture reconstruction quality and that noisy, limited-resolution ground truth may constrain measurable performance gains. Significant challenges remain, all methods exhibit substantially higher errors on vegetation compared to man-made structures, and performance degrades under challenging viewing geometries.
We release the code to enable reproducible use of these methods. in large-scale Earth observation applications.

\small
\bibliographystyle{IEEEtranN}
\bibliography{references}
\end{document}